\documentclass{article}

\usepackage{algorithm}
\usepackage[noend]{algpseudocode}
\usepackage{todonotes}
\usepackage{booktabs}

\usepackage{PRIMEarxiv}

\usepackage[utf8]{inputenc} 
\usepackage[T1]{fontenc}    
\usepackage{hyperref}       
\usepackage{url}            
\usepackage{booktabs}       
\usepackage{amsfonts}       
\usepackage{nicefrac}       
\usepackage{microtype}      
\usepackage{lipsum}
\usepackage{fancyhdr}       
\usepackage{graphicx}       
\graphicspath{{media/}}     

\title{Hybrid Minimax-MCTS and Difficulty Adjustment for General Game Playing
\thanks{
Accepted for publication at the XXII Braziliam Simposium on Games and Digital Entertainment (SBGames 2023)
} 
}

\author{
  Marco Antônio Athayde de Aguiar Vieira \\
  Universidade Federal do Rio Grande do Sul \\
  Porto Alegre, RS, Brasil\\
  \texttt{marco.vieira@inf.ufrgs.br} \\
   \And
  Anderson Rocha Tavares \\
  Universidade Federal do Rio Grande do Sul \\
  Porto Alegre, RS, Brasil\\
  \texttt{artavares@inf.ufrgs.br} \\
   \And
  Renato Perez Ribas \\
  Universidade Federal do Rio Grande do Sul \\
  Porto Alegre, RS, Brasil\\
  \texttt{rpribas@inf.ufrgs.br} \\
}

\begin{document}
\maketitle

\begin{abstract}
Board games are a great source of entertainment for all ages, as they create a competitive and engaging environment, as well as stimulating learning and strategic thinking. It is common for digital versions of board games, as any other type of digital games, to offer the option to select the difficulty of the game. This is usually done by customizing the search parameters of the AI algorithm. 
 However, this approach cannot be extended to General Game Playing agents, as different games might require different parametrization for each difficulty level. In this paper, we present a general approach to implement an artificial intelligence opponent with difficulty levels for zero-sum games, together with a propose of a Minimax-MCTS hybrid algorithm, which combines the minimax search process with GGP aspects of MCTS. This approach was tested in our mobile application LoBoGames, an extensible board games platform, that is intended to have an broad catalog of games, with an emphasis on accessibility: the platform is friendly to visually-impaired users, and is compatible with more than 92\% of Android devices. The tests in this work indicate that both the hybrid Minimax-MCTS and the new difficulty adjustment system are promising GGP approaches that could be expanded in future work.
\end{abstract}


\section{Introduction}
General Game Playing (GGP) is recognized as a significant challenge for Game AI development. Algorithms cannot be tuned in advance for specific games, since they receive declarative descriptions of arbitrary games at run time and must use such descriptions to play those games effectively \cite{Genesereth2005ggp}.  
Variants of Monte Carlo Tree Search (MCTS) are widely used in GGP because MCTS does not depend on game-specific heuristics, building its estimates with samples obtained via rollouts  \cite{monte_carlo}. MCTS builds asymmetric trees, focusing on the moves it considers important.  
However, this asymmetry can lead MCTS to miss crucial moves and fall into tactical trap situations \cite{Baier2014minimax-mcts}. Full-width minimax search does not suffer from this weakness but requires game-specific heuristics to estimate state-values, once the maximum game tree depth is reached due to time constraints  \cite{Knuth1975alphabeta}. 
A combination of the strengths of MCTS and minimax is therefore desired.

Another important issue is that most of the current Game AI research focuses on playing (near-)optimally, often disregarding the aspect of human entertainment that an adjusted difficulty level would provide.  
Difficulty adjustment in GGP is even more challenging, as the algorithm strength must be regulated in runtime, without prior knowledge.  

Other approaches to combine minimax and MCTS tried to use the selective strategy of MCTS together by minimax tactics, which mitigate but fails to eliminate the selectivity related problems \cite{Baier2014minimax-mcts}.  
Furthermore, regarding difficulty adjustment, GGP algorithms usually aim to play (near-)optimally, not allowing strength adjustment. On the other hand, approaches that allow difficulty adjustment are usually a specific algorithm implemented and tuned to a single game, which occurs in most commercial board game apps. 

This work proposes a hybrid Minimax-MCTS algorithm for General Game Playing, together with an algorithm-independent difficulty adjustment mechanism. The hybrid Minimax-MCTS uses minimax with alpha-beta pruning \cite{Knuth1975alphabeta} to build the game tree up to its maximum depth.  
Then, instead of applying a predefined heuristic function to estimate state-values at the maximum depth, we estimate the state value with the average rewards obtained by MCTS's playouts, which play the remainder of the game with random moves. 
We employ an iterative-deepening mechanism, which increases the maximum search depth until the time limit. 
The resulting algorithm combines the full-width tactical awareness of minimax with the domain independence of MCTS, being a GGP algorithm with anytime behavior due the iterative deepening. 

Our difficulty adjustment mechanism uses the game-playing algorithm to estimate the action-values from a given state and Gaussian sampling to select a move. The Gaussian's parameters adjust the desired difficulty level. For instance, lower mean increases the probability of returning actions with lower estimates, resulting in an easier opponent for the human player. 
The difficulty adjustment mechanism can be used with any algorithm that estimates action-values from a given state, which virtually all game-playing algorithms do. 

We evaluate our approach on LoBoGames\footnote{LoBoGames is available at \url{https://play.google.com/store/apps/details?id=com.marcoantonioaav.lobogames}}, a lightweight, accessible, extensible virtual platform for board games, implemented as a mobile app. In the matter of accessibility, the LoBoGames app can fully interact with the device's screen reader, being accessible to visual-impaired users while remaining usable for the non-impaired. The platform currently has five implemented games. 

Our focus is to make it easier to create an engaging and fun environment for the player by creating a unique general system to regulate the difficulty of the games. 
Our results indicate that our hybrid Minimax-MCTS algorithm is a promising GGP algorithm, being able to perform well on a game with long-term reasoning requirements, but falls short on games where initial moves are decisive.
Moreover, our difficulty adjustment mechanism was useful, but requires a competent game-playing algorithm to properly evaluate moves and provide an appropriate level of challenge for human players.  

The contributions of this paper can be summarized as follows: 

\begin{itemize} 
    \item A hybrid Minimax-MCTS algorithm for General Game Playing, which uses MCTS's rollouts rather than a predefined heuristic to estimate state-values at the minimax tree's leaves; 
    \item An algorithm-independent difficulty adjustment mechanism, which receives action-value estimates and selects the returned action according to a Gaussian distribution with user-configured parameters;  
    \item An extensible, lightweight, accessible, mobile application for board games, where the hybrid Minimax-MCTS and the difficulty adjustment mechanism are deployed. The app provides programming interfaces for both the game and the playing algorithm implementation, with accessibility for visually impaired users and compatibility with most Android devices. 
\end{itemize} 

The companion source code of the experiments on the Hybrid Minimax-MCTS algorithm (described on Section \ref{sec:hybrid_evaluation}) is available\footnote{Hybrid Minimax-MCTS experiments source code available at \url{https://github.com/marcoantonioaav/Hybrid-Minimax-MCTS}}. The remainder of this work is organized as follows: Section \ref{sec:background} presents a brief background about the board games properties, AI algorithms and a definition of difficulty levels; Section \ref{sec:related_work} presents other works that implemented Minimax-MCTS hybrids and difficulty adjustment; Section \ref{sec:hybrid} presents our new Minimax-MCTS hybrid algorithm; Section \ref{sec:selection} presents our difficulty adjustment mechanism; Section \ref{sec:exp} presents our mobile game platform and the evaluation of our algorithms; finally, Section \ref{sec:conclusion} presents concluding remarks and directions for future work. 

\section{Background}\label{sec:background} 

On this section, we present the basic concepts that are used along this work, according to the following organization: Section \ref{sec:ggp} introduces the GGP study field and describes the zero-sum games aspects; Section \ref{sec:minimax} presents an overhaul view on AI algorithms applied to zero-sum games and how it behaves on GGP context; Section \ref{sec:ai_difficulty} defines the difficulty levels goals on game AI.

\subsection{General Game Playing and Board Games}
\label{sec:ggp} 
 
General Game Playing (GGP) is a field that focuses on developing intelligent agents capable of playing a wide range of games with little to no prior knowledge or specific domain expertise. 
GGP algorithms since receive a declarative description, or, equivalently, the forward model of arbitrary games at run time and must use such structures to play those potentially unknown games effectively \cite{Genesereth2005ggp}.  

In GGP, games are usually defined by a game description language, with a specific grammar. However, a GGP platform can be built by providing a programming interface that allows an algorithm to query the forward model of the game (i.e., the consequence of performing an action in a state).
This way, any game can be defined, provided that it abides by the programming interface.

In GGP, arbitrary games can be defined; such games may not even be playable for human players by not allowing proper visualization or interaction. 
However, this work focuses on board games, which generally involve the use of pieces and a pre-marked surface, the board.
Such games can be played and enjoyed by humans. In special, our definition of board games does not include other games from the \textit{tabletop} category, which include cards, dice and other elements. 
Moreover, we focus on two-player, combinatorial, zero-sum games. These definitions are explained next.

\subsubsection{Combinatorial zero-sum games} 

Combinatorial games have specific properties over the rest such as the guarantee of determinism and perfect information. 
Conceptually, such games allow the generation of a complete game tree, where nodes are states and edges indicate the transitions that actions perform.
The leaves of the complete game tree are terminal, endgame states, that provide the utility values, or rewards, for each player. 
Game trees present relevant characteristics for the analysis of a game, such as the depth and the width, called \textit{branching factor} (number of possible moves per node).  
A complete game tree allows the inference of the optimal move for all possible states. 
In practice, the exponential complexity of the game tree limits the ability to explore it in a satisfactory amount of time. Therefore, artificial intelligence algorithms often partially explore it.  

\begin{figure}[H] 
    \centering    
        \includegraphics[width=0.6\linewidth]{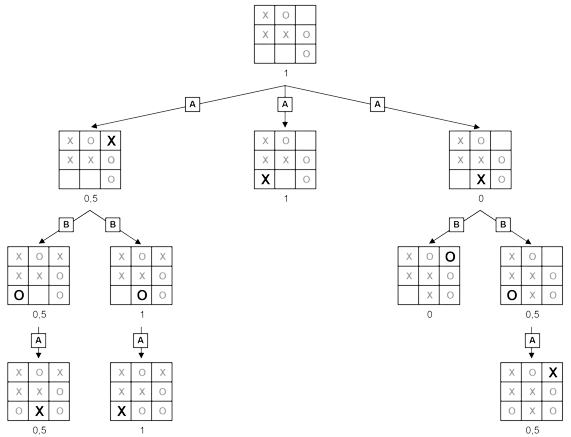} 
        \caption{A part of the tic-tac-toe's game tree. Image by Melk{\'o} and Nagy \cite{game_tree}.} 
    \label{fig:game_tree} 
\end{figure}

In zero-sum games, the victory of one player results in the defeat of the other. In other words, the rewards on terminal states are opposite. 
This way, we only need to represent the rewards for a player, which tries to maximize it whereas the opponent tries to minimize it. 
Such trees are called minimax trees.
Figure \ref{fig:game_tree} illustrates a combinatorial zero-sum game tree, by showing a part of tic-tac-toe's tree, where ``A'' symbolizes an action of player X and ``B'' symbolizes an action of player O. 

\subsection{Zero-sum Game AI} 
\label{sec:minimax} 
The pursued ideal of most artificial intelligence algorithms in zero-sum games is to explore the game tree and search for a path that is guaranteed to lead to a win. 
In most cases, the entire tree cannot be explored due to its exponential complexity. Thus, most AI algorithms use different strategies to partially explore the tree and still gather enough information to identify a decent move. 

The minimax algorithm is one of the classic artificial intelligence algorithms for board games \cite{Knuth1975alphabeta}. 
It explores the definition of zero-sum games, a player is considered to be a reward maximizer and its opponent is the reward minimizer. 
The search explores the game tree depth-wise until it finds a terminal state of the game (leaf node of the game tree), where the utility of the state is evaluated according to the win conditions of the game itself: maximum value for a victory, minimum value for a defeat and intermediate value for a draw. 

\begin{figure}[H] 
     \centering 
         \includegraphics[width=\linewidth]{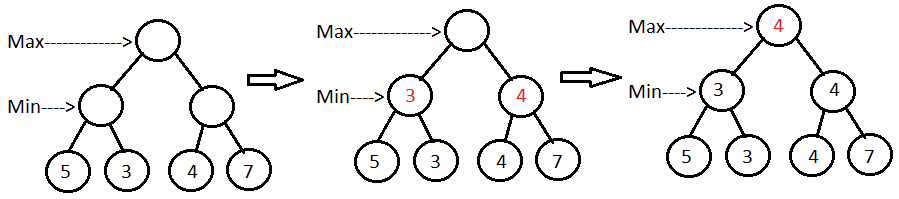} 
         \caption{Example of execution of the minimax algorithm. Image by Baeldung (\url{https://www.baeldung.com/java-minimax-algorithm}).} 
     \label{fig:minimax} 
\end{figure}

Minimax algorithm infers the value of the non-terminal states by using the maximum value of successor nodes if the state belongs to the maximizing player, or the minimum value of the successors otherwise. The values are calculated backwards until the root node. 
Figure \ref{fig:minimax} illustrates the algorithm behavior, with labels identifying the nodes of the max (maximizer) and min (minimizer) players. 

Minimax is usually implemented with alpha-beta pruning \cite{Knuth1975alphabeta}, which discards moves with provably worse values than alternative moves already explored. 
Moreover, limited tree depth is employed to return a move in a reasonable time. 
In non-terminal states at maximum exploration depth, a heuristic evaluation function is used to estimate the utility of the state obtained by the agent. 
The elaboration of this heuristic usually requires the programmer's previous knowledge about the game to be played by the agent, which limits this algorithm, making it efficient only for a pre-established number of games. 

The domain-specific nature of minimax limits its application in General Game Playing domains. 
Some algorithms stand out to be more suitable for building GGPs than minimax, such Monte Carlo Tree Search (MCTS) \cite{monte_carlo} or Sequential Halving On Trees (SHOT) \cite{cazenave2014sequential}. 
These algorithms use random game simulations, also called \textit{rollouts} or \textit{playouts}, to generate state-value estimates. 
MCTS has been more successful than SHOT on GGP \cite{Genesereth2005ggp} and we focus our discussion on MCTS thereafter. 
MCTS grows the game tree towards nodes it regards as promising, leaving few opportunities to probe the state-values on other regions.
MCTS is thus domain-agnostic and grows the game tree asymmetrically. 
However, this asymmetry and the low number of estimates on so-called unpromising regions do not allow MCTS to prove the value of a node. 
This has been shown to lead MCTS to miss crucial moves and fall into tactical trap situations \cite{Baier2014minimax-mcts}.

\subsection{Game AI with difficulty levels} 
\label{sec:ai_difficulty}
  
It is expected that a game for human enjoyment must have the ability to provide a fun experience for the player. To achieve that, the game cannot be so hard that it is frustrating or so easy that it is boring \cite{weber2020dynamic}. This statement implies on a subjective balance problem, but it is possible to assume that, in general, the following rule must be followed: a level of difficulty cannot make the agent invincible and neither incapable of winning.  

A basic assumption on the topic of difficulty levels on games is very simple: the greater the difficulty level, the better that the agent should play the game. This assumption also applies to the other way, meaning that the agent should also play worse on lower difficulty levels.  

Based on this assumption, the difference in the quality of gameplay on each level should be clear. The optimal distance between each level is also a subjective problem, but it is desirable that this distance can be perceived by a human player so that the performance of the agent can be balanced in the best way that is intended by the authors.  

\section{Related work}\label{sec:related_work} 

\subsection{MCTS-Minimax Hybrids} 

The concept of combining minimax with MCTS strategies is not an innovation. The paper of Hendrik Baier and Mark HM Winands \cite{Baier2014minimax-mcts} has proposed three approaches to that concept: Using minimax in the selection/expansion phase, the rollout phase, and the backpropagation phase of MCTS. 

There are many possible benefits of this combinations, such as: 

\begin{itemize} 
    \item Save expansion processing in minimax; 
    \item Increase minimax's max depth; 
    \item Avoid traps in MCTS; 
    \item Avoid shallow losses in MCTS. 
\end{itemize} 

What differs the most from our approach is that these uses MCTS framework, while our tactic is to use the minimax search as the basis of the algorithm. The main reason behind our algorithm been hybrid is to adapt the minimax decision process into a GGP scenario, so the selection policy present in MCTS is not used. 

\subsection{Difficulty adjustment} 

Some implementations of difficulty levels try to dynamically adapt the difficulty through the player feedback. For example, the AdaptiveMiniMax (AMM) algorithm proposes an extension to the minimax algorithm (described in Section \ref{sec:minimax}) where the quality of the selected moves is adapted according to the average evaluation of the plays of the human opponent \cite{dziedzic2016dynamic_board}. Another property that is also considered in the article in which this algorithm is proposed is the depth of exploration of the AMM.  

On the other hand, some other implementations define the difficulty levels as fixed sets of the agent's parameter valuations. The open source Lichess Chess platform\footnote{Available at \url{https://lichess.org/}} uses the Stockfish evaluation engine, which in turn implements a version of the minimax algorithm. On the open code, it is possible to analyze the three parameters used to differentiate the eight levels of difficulty available on the platform: Duration, Skill Level and depth. According to the documentation of the Fairy Stockfish engine \cite{fairy_stockfish}, the Skill Level parameter determines the chance of selecting a move other than the best evaluated one, without further explanation of how this selection is made. 

Learning from these implementations, we identified a pattern of two types of properties that are most considered: Choice’s reward (agent’s performance) and choice’s complexity (problem’s difficulty). A choice-reward driven agent uses a policy that selects a move based on its evaluation, such as purposely choosing a low evaluated move to simulate a low level of difficulty. On the other hand, an agent that uses the complexity of the choices varies the limits of its evaluation through its exploration parameters, such as the maximum depth or execution time.  

\section{Hybrid Minimax-MCTS for General Game Playing}\label{sec:hybrid} 
\label{sec:minimaxMCTS}

The game-playing algorithm is a crucial part of our approach. In special, we need an algorithm to evaluate and rank the moves by their quality or utility. In this paper, we convention that moves' values are within the range $[0, 1]$, with the following semantics: 

\begin{itemize} 
     \item The evaluation value $0$ characterizes a move that leads to the defeat of the agent, if the opponent exploits it correctly; 
     \item The evaluation value $0.5$ characterizes a move that leads to a draw (if there is a draw state), or even that prolongs the game without contributing to the change of advantage; 
     \item The evaluation value $1$ characterizes a move that leads to the win of the agent, which the opponent has no means to avoid. 
\end{itemize} 

To perform move evaluation we propose a hybrid Minimax-MCTS algorithm, which combines the tactical advantages of minimax with the GGP elements of MCTS.  

The basis of the AI algorithm is the minimax search with alpha-beta pruning. Minimax is able to clearly identify imminent victories and defeats. Besides, as minimax does not have any policy that prioritizes the exploration of promising nodes, this guarantees that all moves can be explored in the same depth. This is an important feature as the intent behind this step is not just to figure out the best move, but to evaluate every move as accurately as possible. 

The only problem with original minimax  algorithm in this scenario is the dependency of previous knowledge about the game, needed to elaborate a heuristic function to evaluate non-terminal states. Hence, we incorporate the concept of playouts into the minimax algorithm to solve this problem, making the evaluation process more general and turning the agent into a GGP agent (Algorithm \ref{alg:hybrid}).
We replace the heuristic evaluation with the mean of a fixed amount of $P$ playouts.
These playouts are random simulations used to estimate the utility of a given game state. Usually the playout aims to simulate the remainder of the game up to its end, but we add a depth limit to avoid possible infinite loops caused by repeated states.  

\begin{algorithm} 
\caption{Hybrid Minimax-MCTS with alpha-beta pruning}\label{alg:hybrid} 
\begin{algorithmic}[1] 
\Procedure{Minimax-MCTS}{$state$, $depth$, $\alpha$, $\beta$} 
\If{$state$ is terminal} 
    \State \Return evaluation of $state$ 
\EndIf 
\If{$depth$ = 0} 
    \State \Return $\frac{P \: playouts \: from \: state}{P}$ 
\EndIf 
\If{is maximizer's turn} 
    \State $u \gets -\infty$  
    \ForAll{possible $moves$ in $state$} 
        \State $nextState \gets$ apply $move$ in $state$ 
        \State $v \gets$ Minimax-MCTS($nextState$, $depth-1$, $\alpha$, $\beta$) 
        \State $u \gets max(u, v)$  
        \If{$u \geq \beta$} 
            \State $break$ 
        \EndIf 
        \State $\alpha \gets max(u, \alpha)$ 
    \EndFor 
    \State \Return $u$ 
\Else 
    \State $u \gets \infty$  
    \ForAll{possible $moves$ in $state$} 
        \State $nextState \gets$ apply $move$ in $state$ 
        \State $v \gets$ Minimax-MCTS($nextState$, $depth-1$, $\alpha$, $\beta$) 
        \State $u \gets min(u, v)$  
        \If{$u \leq \alpha$} 
            \State $break$ 
        \EndIf 
        \State $\beta \gets min(u, \beta)$ 
    \EndFor 
    \State \Return $u$ 
\EndIf 
\EndProcedure 
\end{algorithmic} 
\end{algorithm} 

Finally, the depth of the minimax search isn't predefined. As mentioned before, the complexity of the games is not previously known, so it is hard to define a fixed depth value that is going to deliver satisfactory results in a constant amount of time for all games. 
Hence, we use iterative deepening (Algorithm \ref{alg:iterative}), so that the algorithm uses all the given time budget. 

\begin{algorithm} 
\caption{Iterative deepening evaluation}\label{alg:iterative} 
\begin{algorithmic}[1] 
\Procedure{Iterative Deepening}{$state$, $time$} 
\State $d \gets 0$ 
\While {$time$ not reached} 
    \ForAll{possible $moves$ in $state$} 
        \State $s' \gets$ apply $move$ in $state$ 
        \State $evaluation[move] \gets$ Minimax-MCTS($s'$, $d$, $0$, $1$) 
    \EndFor 
    \State $d \gets d + 1$ 
\EndWhile 
\State \Return $evaluation$ 
\EndProcedure 
\end{algorithmic} 
\end{algorithm} 

\section{GGP difficulty-driven move selection}\label{sec:selection} 
\label{sec:difficulty}
On the move selection algorithm, the inputs are strictly the evaluated available states and the difficulty level. There is no access to any information about the game or about the evaluation process.  

This adjustment algorithm (Algorithm \ref{alg:selection}) applies a stochastic selection, to have variations in the chosen moves. This algorithm starts by sampling a value, and then selects the move with the closest evaluation to this value. The idea of balancing is that in more difficult levels the sampled value tends to be higher than in easier levels, implying the selection of moves with better evaluation. 

The value is sampled from a Gaussian distribution, because in this way we can use the mean $\mu$ and the standard deviation $\sigma$ of the distribution as the parameters that define each difficulty. Balancing the performance of the agents should be a game design choice, but care must be taken that the valuations of $\mu$ and $\sigma$ do not eliminate the possibility that the agent chooses winning or losing moves.  

\begin{algorithm} 
\caption{Move selection algorithm}\label{alg:selection} 
\begin{algorithmic}[1] 
\Procedure{Stochastic Selection}{$evaluation$, $\mu$, $\sigma$} 
\State $target \sim \mathcal{N}(\mu, \sigma)$ \Comment{sampling from normal (Gaussian)} 
\State $difference \gets \infty$ 
\ForAll{possible $moves$ in $evaluation$} 
    \If{$\left | evaluation[move] - target \right | < difference$} 
        \State $selected \gets move$ 
        \State $difference \gets \left | evaluation[move] - target \right |$ 
    \EndIf 
\EndFor 
\State \Return $selected$ 
\EndProcedure 
\end{algorithmic} 
\end{algorithm} 

\section{Experiments} 
\label{sec:exp} 

\subsection{Evaluation Platform: LoBoGames} 
\label{sec:lobogames} 

Our evaluation testbed is an accessible virtual board game platform, implemented as a Android mobile application called LoBoGames. The main feature of this platform is accessibility, promoted mainly through adaptability for visually impaired users. The app was made to be an extensible platform, meaning that it provides interfaces to simplify the process of adding new games. The app is compatible with a large number of devices (Approximately 92.76\% of Android devices, according to Google Play Console). The minimum OS version required is Android 4.1 Jelly Bean, released in June 2012. Our app prototype is available to download for free at the Play Store. 

\subsubsection{Implemented games}  

There are currently 5 games implemented in the app: Tic-tac-toe, Tapatan, Alquerque, Tsoro Yematatu and Five Field Kono. They can be all classified as combinatorial zero-sum games. A short description of each of them follows:

\begin{itemize} 
     \item Tic-tac-toe: Insertion game in a $3\times3$ board. The board starts empty. The goal is to align 3 pieces;
     \item Tapatan: Uses a $3\times3$ board with line connections. The board starts with all 6 pieces (3 pieces of each player) on the game defined start position. You can move your pieces to adjacent connected positions. The goal is to align 3 pieces;
     \item Alquerque: Uses a $5\times5$ board with line connections. The board starts with all 24 pieces (12 pieces of each player) on the game defined start position. You can move your pieces to adjacent connected positions and capture opponent pieces by jumping hover. Multiple captures in a single turn are allowed. It is required to do a capture when possible. The goal is to capture all opponent pieces;
     \item Tsoro Yematatu: Uses a $5\times5$ board with line connections. The board starts empty. You can first position 4 pieces at the board then move your pieces to adjacent connected positions. The goal is to align the 4 pieces;
     \item Five Field Kono: Uses a $5\times5$ board with line connections. The board starts with all 14 pieces (7 pieces of each player) on the game defined start position. You can move your pieces to adjacent connected positions. The goal is to fill the opponent's side of the board with pieces.
\end{itemize} 

\subsubsection{Agent's design}  

The LoBoGames app was designed to have an extendable catalog of games, therefore, the AI agent must handle an indefinite number of games. 

\begin{figure}[H] 
    \centering    
        \includegraphics[width=0.75\linewidth]{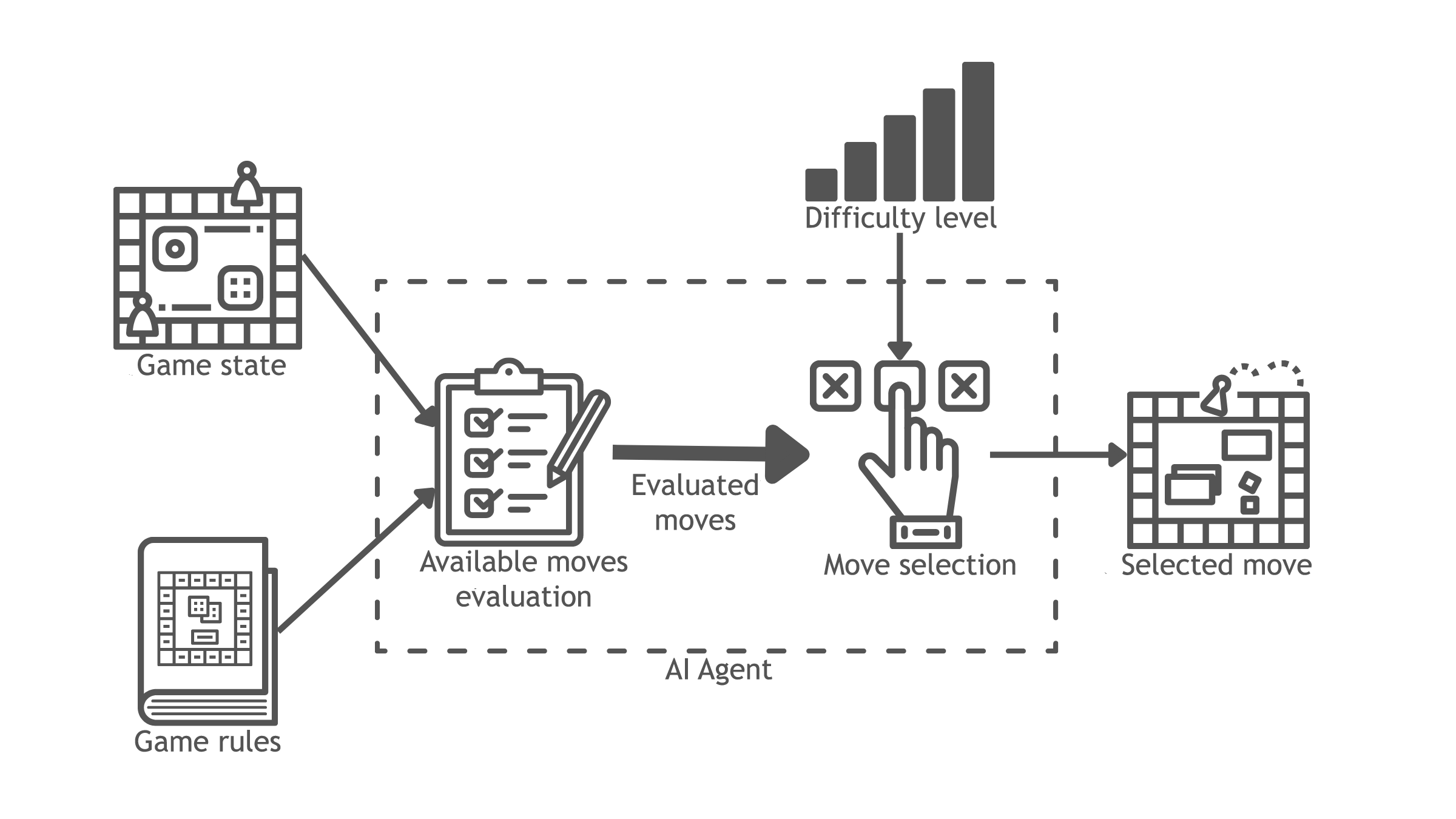} 
        \caption{Agent's decision process. The AI agent performs two independent tasks: move evaluation, performed by the hybrid Minimax-MCTS algorithm (Section \ref{sec:minimaxMCTS}) and move selection with adjustable difficulty (Section \ref{sec:difficulty}). } 
    \label{fig:agent} 
\end{figure}

The agent's play is divided into two stages: Evaluation and Selection. On the evaluation stage, without knowing the level of difficulty, the agent must analyze the moves allowed from a given state and assign a numerical evaluation to each one of them, symbolizing the estimated reward of each choice. 
On the selection stage, the agent chooses the ideal move based only on the selected difficulty level and the evaluation carried out by the previous step. Figure \ref{fig:agent} illustrates the flow of the agent's decision process. 

\subsubsection{Difficulty levels}  

Our app was defined as containing 3 different difficulty levels: Easy, Medium and Hard. Table \ref{tab:difficulty1} shows the LoBoGames app agent difficulty parameters of the approach depicted in Section \ref{sec:difficulty}. The difficulty level can be selected as an option at the beginning of a single-player match. 
These values were chosen in previous experiments to allow the desired behavior that, on ``Easy'', the agent usually blunders, but can eventually win a game. On the other hand, on ``Hard'', the agent avoids blunders and is likely to play winning moves if they're found by the evaluation step previously performed (see Section \ref{sec:minimaxMCTS}). 

\begin{table}[ht] 
    \caption{LoBoGames app difficulty levels and corresponding parameters of the move selection mechanism of Section \ref{sec:difficulty}. The resulting Gaussians are illustrated in Figure \ref{fig:gaussian}.} 
    \centering 
        \begin{tabular}{ |c|c|c|  } 
            \hline 
            Level & $\mu$ & $\sigma$\\ 
            \hline 
            Easy & 0.4 & 0.3 \\ 
            Medium & 0.6 & 0.3 \\ 
            Hard & 1.0 & 0.3 \\ 
            \hline 
        \end{tabular} 
    
    \label{tab:difficulty1} 
\end{table} 

\begin{figure}[H] 
     \centering 
         \includegraphics[width=0.6\linewidth]{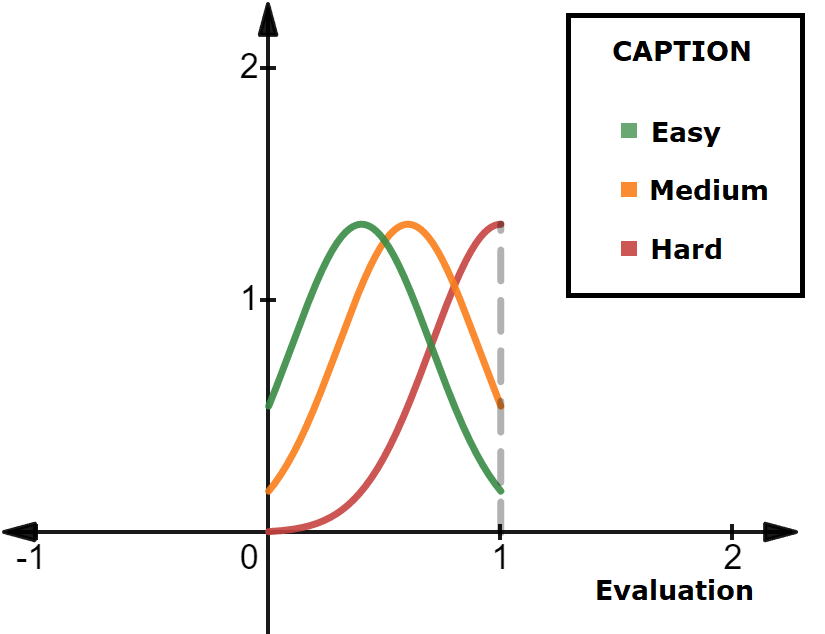} 
         \caption{Distribution of Gaussian distribution for move selection according to each difficulty level on the LoBoGames app.  The Gaussians are defined according to the parameters of Table \ref{tab:difficulty1}.} 
     \label{fig:gaussian} 
\end{figure}

The Gaussian distribution that is defined at each difficulty level can be used to design the agent's behavior, as shown in Figure \ref{fig:gaussian}. Note that values outside the range $[0, 1]$ are clipped to remain in the range. 

Table \ref{tab:difficulty2} presents the probability of choosing a move according to each evaluation (X) and difficulty level, induced by the Gaussians with parameters of Table \ref{tab:difficulty1}. 

\begin{table}[ht] 
    \caption{Probability of choosing a move according to three percentiles of the evaluation (X), for each difficulty level in the LoBoGames' app move selection mechanism. Each difficulty level is a Gaussian, illustrated in Fig. \ref{fig:gaussian}, whose parameters are shown in Tab. \ref{tab:difficulty1}.} 
    \centering 
        \begin{tabular}{ |c|c|c|c|  } 
            \hline 
            Level & $P(X<0.25)$ & $P(0.25\leq X\leq0.75)$ & $P(X>0.75)$\\ 
            \hline 
            Easy & 30.85\% & 56.98\% & 12.17\% \\ 
            Medium & 12.17\% & 56.98\% & 30.85\% \\ 
            Hard & 00.62\% & 19.61\% & 79.77\% \\ 
            \hline 
        \end{tabular} 
    \label{tab:difficulty2} 
\end{table}

\subsection{Hybrid Minimax-MCTS Evaluation}
\label{sec:hybrid_evaluation}

These experiments aim to validate the hybrid algorithm proposed on Section \ref{sec:hybrid} as a viable option for GGP, demonstrating that the proposed algorithm can achieve competitive results when against UCT (MCTS with upper confidence bounds criteria for node selection \cite{monte_carlo}), a common GGP baseline. 

\subsubsection{Experiment details}

The platform chosen to run the tests was the general game system Ludii \cite{piette2019ludii}, a GGP tool built for game archaeology and AI research. This tool provides a large catalog of games, including some of the most famous board games. It also provides an extensible AI environment, with various algorithms implemented, including UCT.

We implemented our hybrid Minimax-MCTS algorithm (Section \ref{sec:minimaxMCTS}) as an extension of the Ludii's AI base class. 
When playing a game, agents have five seconds to return a move. The time limit on Algorithm \ref{alg:iterative} was estimated linearly according to the latest two iterations, so the mean spent time per turn could be closer to the five seconds limit. The amount of playouts ($P$ on Algorithm \ref{alg:hybrid}) per evaluation of the hybrid algorithm was set to 15, and a max playout depth value was set to 100.

We tested our algorithm against the UCT baseline on 4 Ludii games: Tic-tac-toe, Tapatan, Alquerque and Reversi. 
The first three games are also present on LoBoGames, but Reversi is not. Maintaining the app's accessibility for visually impaired users on a larger game such as Reversi is an ongoing work. 

Reversi uses a $8\times8$ board initialized with two pieces for each player in the center.
A player can place a piece on an empty position next to an opponent's piece, a long as a straight line can be drawn between the new piece and another player piece with opponent's pieces in between. 
These opponent pieces are captured by the player.
Players alternate such moves until the board is full and the player with more pieces is the winner.
We do evaluations on Reversi because it is an interesting benchmark, since it requires long-term reasoning (greedily aiming for pieces is bad strategy) and a game always finishes.

Each game is played 20 times, alternating the first and second players. Each playthrough is declared a draw after 100 turns, since some games could cycle indefinitely. 
The three games that are present on both on LoBoGames and Ludii are commented regarding their variations below:

\begin{itemize} 
     \item Tic-tac-toe: Same rules on both platforms;
     \item Tapatan: on LoBoGames, the initial position contains all player pieces. On Ludii, the board starts empty, so the 3 first moves of each player are insertions;
     \item Alquerque: on LoBoGames, multiple captures in a single turn are allowed and captures are forced moves. On Ludii, multiple captures are not allowed and captures are optional.
\end{itemize} 

There are versions of Tsoro and Five Field Kono on Ludii, but their rules differ greatly from the LoBoGames version, as the board design is different. Moreover, the four selected games already cover our evaluation needs, as there are variability on game mechanics (insertion, movement, alignment, capture) and strategic depth (e.g. simple in tic-tac-toe and relatively complex in Reversi).

\subsubsection{Results}

The experiment results (Table \ref{tab:hybridtests}) shows that our algorithm is competitive against UCT. 

\begin{table}[ht] 
    \caption{Games won by our hybrid Minimax-MCTS algorithm, by UCT and drawn on each game.} 
    \centering 
        \begin{tabular}{ cccc  } 
            \toprule
            Game & Hybrid & UCT & Draw\\ 
            \midrule
            Tic-tac-toe & 0\% & 0\% & 100\% \\ 
            Tapatan & 30\% & 50\% & 20\% \\ 
            Alquerque & 0\% & 30\% & 70\% \\ 
            Reversi & 55\% & 40\% & 5\% \\ 
            \bottomrule
        \end{tabular} 
    \label{tab:hybridtests} 
\end{table} 

Both algorithms always draw on tic-tac-toe, suggesting perfect play on such a simple game. 
However, in Tapatan and Alquerque UCT went better.
Ludii's version of Tapatan benefits the first player and only UCT exploited that, winning all games as the first player. 
Other combinations of the hybrid algorithm's parameters (the amount of playouts per evaluation and the maximum playout depth) could improve the performance, as these parameters directly impact the time usage and could change the maximum depth reached by the iterative deepening mechanism.

Alquerque is a much more complex game than the previous two. Our hybrid algorithm seems unable to obtain valuable information on the beginning of the game, playing almost randomly. Even so, our algorithm managed to draw 70\% of the games by reaching a cycle on their final stages. 

On Reversi, our hybrid algorithm outperformed UCT, suggesting the benefits of its tactical awareness. 
UCT grows its search tree asymmetrically towards states it deems as promising, i.e., the search tree is deeper towards the ``promising'' states and shallower on ``unpromising'' ones.
This might make the algorithm fail to differentiate good and bad lines of play on states it deems as unpromising. 
The minimax part of the hybrid algorithm performs a full-width search and expands all states on the limit depth.
Hence, it might be able to differentiate good and bad situations on the limit-depth states, performing moves that UCT might miss.

Our hybrid Minimax-MCTS algorithm is a promising GGP approach, since it seemed to exhibit tactical awareness on the strategically deeper game of Reversi. However, it needs enhancements to outperform the UCT baseline on games such as Tapatan and Alquerque. 

\subsection{Difficulty Adjustment Evaluation} 

The application was tested by more than 30 users in the Play Store, and in addition to that the app was tested with a group of three blind users. The reports were collected in a unstructured interview format. 

The reports about the difficulty levels were majorly positive for most of the games. In the games Tsoro Yematatu and Five Field Kono, the difficulty levels were identified as ``easier than expected'' by some users, even in the Hard level. 

The problem that happens with these games is that, due to their complexity aspects, such as branching factor and terminal nodes depth, the iterative deepening of the agent does not reach the required depth to give better results in the time that was reserved by the app. Additionally, playouts appear to be unable to obtain information upon reaching a terminal state due to the cycles and depth of the terminals.

\section{Conclusion and Future Work}\label{sec:conclusion} 

Game playing algorithms are either precise, but game-specific (e.g. minimax) or game-agnostic but imprecise, falling into tactical traps due to missing good moves due to the randomness in the information acquisition mechanism (e.g. MCTS).
Moreover, most of the Game AI research focuses on (near-)perfect play, often disregarding the aspect of human entertainment that a proper difficulty level would provide.  

This work proposes a hybrid Minimax-MCTS algorithm combining the best of minimax and MCTS.
Our algorithm uses minimax with alpha-beta pruning \cite{Knuth1975alphabeta} to build a full-width game tree up to a depth limit.  
Then, instead of applying a predefined heuristic function to estimate state-values at the limit depth, we estimate state-values with the average rewards obtained by heuristic-free MCTS's playouts.
An iterative-deepening mechanism enables the full use of any allotted time for our algorithm.
The resulting algorithm is a general game player due to MCTS's heuristic-free playouts with the tactical awareness of minimax due to its full-width game tree construction. Moreover, the algorithm is anytime due to the iterative deepening. 

Our difficulty adjustment mechanism uses a game-playing algorithm as a subroutine to estimate the action-values from a given state. Then, Gaussian sampling is applied to select a move. The Gaussian's parameters adjust the desired difficulty level. 
The difficulty adjustment mechanism can be plugged into any game-playing algorithm that estimates action-values, which include virtually all game-playing algorithms. 

The novel difficulty adjustment mechanism can be used to generate a full set of difficulty levels with distinct behavior and is not limited by the number of levels, the target game or the game-playing algorithm that estimates values. The balancing task showed to be simple, since the difficulty parameters impact can be analyzed just by calculating the Gaussian distribution itself. 

Our evaluation platform is the LoBoGames app, a lightweight and extensible virtual platform for board games. 
It has emphasis on accessibility: the app can fully interact with the device's screen reader, being accessible to visual-impaired users while remaining usable for the non-impaired. Moreover, the app has minimum requirements, being compatible with more than 92\% of Android devices. 

Our hybrid Minimax-MCTS algorithm is a promising GGP approach, since it seemed to exhibit tactical awareness on strategically deeper games. However, it needs enhancements to outperform the UCT baseline on games where the initial moves are decisive.
Such enhancements could allow a better assessment of our difficulty adjustment mechanism, since some users reported that, even the Hard level was unable to be challenging.

Future work could try to add the evaluation search complexity aspects as difficulty parameters as well. That would unite the concepts of choice-reward and choice-complexity, allowing to distinguish the complexity of each move too instead than just controlling the probability of choosing the right move. Literature suggests that the complexity of the move depends mainly on the depth of the game tree \cite{dziedzic2016dynamic_board}. 

On the other hand, the hybrid Minimax-MCTS algorithm didn't perform well when trying to play games with complex tree aspects, such as high branching factor and repeated states (transpositions). Some ideas to reduce this effect are: 

\begin{itemize} 
     \item Implement the algorithm with transposition tables; 
     \item Add other types of pruning than alpha-beta, (e.g. probabilistic pre-pruning without probing the node's value \cite{Buro1997multiprobcut}); 
     \item Mix with other hybrid Minimax-MCTS algorithms techniques (e.g. \cite{Baier2014minimax-mcts}). 
\end{itemize} 

\section*{Acknowledgments}
This study was financed in part by the Coordenação de Aperfeiçoamento de Pessoal de Nível Superior - Brasil (CAPES) - Finance Code 001.

This work was also partially funded by Conselho Nacional de Desenvolvimento Científico e Tecnológico (CNPq).

\bibliographystyle{unsrt}  
\bibliography{references}

\begin{thebibliography}{10}

\bibitem{Genesereth2005ggp}
Michael Genesereth, Nathaniel Love, and Barney Pell.
\newblock General game playing: Overview of the aaai competition.
\newblock {\em AI magazine}, 26(2):62--62, 2005.

\bibitem{monte_carlo}
Cameron~B Browne, Edward Powley, Daniel Whitehouse, Simon~M Lucas, Peter~I Cowling, Philipp Rohlfshagen, Stephen Tavener, Diego Perez, Spyridon Samothrakis, and Simon Colton.
\newblock A survey of monte carlo tree search methods.
\newblock {\em IEEE Transactions on Computational Intelligence and AI in games}, 4(1):1--43, 2012.

\bibitem{Baier2014minimax-mcts}
Hendrik Baier and Mark~HM Winands.
\newblock Mcts-minimax hybrids.
\newblock {\em IEEE Transactions on Computational Intelligence and AI in Games}, 7(2):167--179, 2014.

\bibitem{Knuth1975alphabeta}
Donald~E Knuth and Ronald~W Moore.
\newblock An analysis of alpha-beta pruning.
\newblock {\em Artificial intelligence}, 6(4):293--326, 1975.

\bibitem{game_tree}
Ervin Melk{\'o} and Benedek Nagy.
\newblock Optimal strategy in games with chance nodes.
\newblock {\em Acta Cybernetica}, 18(2):171--192, 2007.

\bibitem{cazenave2014sequential}
Tristan Cazenave.
\newblock Sequential halving applied to trees.
\newblock {\em IEEE Transactions on Computational Intelligence and AI in Games}, 7(1):102--105, 2014.

\bibitem{weber2020dynamic}
Matheus Weber and Pollyana Notargiacomo.
\newblock Dynamic difficulty adjustment in digital games using genetic algorithms.
\newblock In {\em 2020 19th Brazilian Symposium on Computer Games and Digital Entertainment (SBGames)}, pages 62--70. IEEE, 2020.

\bibitem{dziedzic2016dynamic_board}
Dagmara Dziedzic.
\newblock Dynamic difficulty adjustment systems for various game genres.
\newblock {\em Homo Ludens}, 9(1):35--51, 2016.

\bibitem{fairy_stockfish}
Fairy-Stockfish.
\newblock Fairy-stockfish, 2023.

\bibitem{piette2019ludii}
Eric Piette, Dennis~JNJ Soemers, Matthew Stephenson, Chiara~F Sironi, Mark~HM Winands, and Cameron Browne.
\newblock Ludii--the ludemic general game system.
\newblock {\em arXiv preprint arXiv:1905.05013}, 2019.

\bibitem{Buro1997multiprobcut}
Michael Buro.
\newblock {Experiments with Multi-ProbCut and a new high-quality evaluation function for Othello}.
\newblock {\em Games in AI Research}, 34(4):77--96, 1997.

\end{thebibliography}

\end{document}